\documentclass{article}

\usepackage{arxiv}
\usepackage[utf8]{inputenc} % allow utf-8 input
\usepackage[T1]{fontenc}    % use 8-bit T1 fonts
\usepackage{hyperref}       % hyperlinks
\usepackage{url}            % simple URL typesetting
\usepackage{booktabs}       % professional-quality tables
\usepackage{amsfonts}       % blackboard math symbols
\usepackage{amsmath}
\usepackage{nicefrac}       % compact symbols for 1/2, etc.
\usepackage{microtype}      % microtypography
\usepackage{graphicx}
\usepackage{natbib}
\usepackage{doi}
\usepackage{enumitem}

\title{OrcaRouter: A Production-Oriented LLM Router \\
       with Hybrid Offline--Online Learning}

\author{%
  Zhenghua Bao \quad Fengya Tian \quad Chris Zhang \quad Zhenjun Chen \quad Xile Ma \quad Yi Shi$^{\dagger}$ \\[0.5em]
  Continuum AI \\
  \texttt{hello@continuum01.ai} \\[0.25em]
  {\small $^{\dagger}$Project lead and corresponding author}
}

\date{}

\renewcommand{\headeright}{Technical Report}
\renewcommand{\undertitle}{}
\renewcommand{\shorttitle}{OrcaRouter}

\hypersetup{
  pdftitle={OrcaRouter: A Production-Oriented LLM Router with Hybrid Offline--Online Learning},
  pdfsubject={cs.LG; cs.CL; cs.AI},
  pdfauthor={Zhenghua Bao, Fengya Tian, Chris Zhang, Zhenjun Chen, Xile Ma, Yi Shi},
  pdfkeywords={LLM routing, contextual bandits, LinUCB, hybrid offline-online learning, cost-aware LLM inference},
}

\begin{document}
\maketitle

\begin{abstract}
The rapid development of large language models, each with distinct
capabilities and inference costs, raises a practical deployment
question: given an incoming request, which model should handle it?
We present \textbf{OrcaRouter}\footnote{\url{https://www.orcarouter.ai/}},
a production-oriented LLM router
that combines a LinUCB-based contextual bandit over lexical and
sentence-embedding features with a hybrid offline--online learning
protocol. Offline, OrcaRouter obtains full-information feedback by evaluating
each candidate model on a curated set of routing prompts, yielding a
reward matrix used to fit one ridge regressor per arm. At deployment time, it initializes from these
parameters and can optionally continue learning from bandit feedback,
updating only the selected model's arm after observing its reward.
At the time of our RouterArena submission (May 20, 2026),
OrcaRouter-Adaptive ranked second on the public RouterArena leaderboard
with an arena score of $\mathbf{72.08}$, achieving $75.54\%$ accuracy
at \$1.00 per 1K queries.
\end{abstract}


\section{Introduction}

The ecosystem of Large Language Models (LLMs) has expanded rapidly
with the scaling of autoregressive models such as
GPT-3~\citep{brown2020language}. Today, deployment systems can choose
from a large and rapidly changing pool of models that differ
substantially in capability, latency, and inference cost. For any given
request, this choice directly affects both response quality and
operating cost. Yet in many production systems, model selection is
still handled by coarse heuristics or by defaulting to a single strong,
expensive general-purpose model, regardless of the request's task type,
complexity, or cost sensitivity.

\paragraph{The production problem.}
The task of an LLM router is to select, for each request, the model
that offers the best expected quality--cost trade-off. This differs
from reward modeling~\citep{ouyang2022training}, where a learned
reward model scores generated outputs after candidate responses have
already been produced. In routing, the decision must be made before
generation, when only the request and model metadata are available.

Existing work approaches cost-aware LLM routing from several
perspectives. \citet{shnitzer2023large} study the use of benchmark datasets for
training routers that select among candidate LLMs. FrugalGPT~\citep{chen2023frugalgpt} uses an LLM cascade that learns which
combinations of models to invoke for different queries, often
escalating from cheaper to stronger models when needed.
RouteLLM~\citep{ong2024routellm} learns preference-data routers that primarily choose between a
stronger and a weaker model at inference time.
OrcaRouter instead formulates routing as a configurable multi-arm
contextual bandit, allowing the model pool to change over time and the
routing policy to adapt from deployment feedback.

For a production-oriented router, three requirements are central:
(i) it should support a configurable and frequently changing model
pool, so that operators can incorporate newly released models;
(ii) it should adapt to shifting request distributions; and
(iii) it should support continual adaptation, refining its routing
policy after each served request based on the feedback observed in
that round.

\paragraph{Contributions.}
Rather than proposing a new bandit algorithm, this technical report
studies how contextual bandits can be applied to production-oriented
LLM routing with evolving model pools, full-information offline warmup,
and partial-information online adaptation.

\begin{enumerate}[leftmargin=*, itemsep=2pt]
  \item We formulate LLM routing as a multi-armed contextual-bandit
        problem and implement a LinUCB-based router over lexical and
        sentence-embedding features, with one ridge regressor per
        candidate model.

  \item We introduce a hybrid offline--online learning protocol. The
        router is initialized offline from a full-information reward
        matrix constructed over a curated set of routing prompts. In
        deployment, it can continue updating from bandit feedback after
        each served request.

  \item At the time of our RouterArena submission,
      OrcaRouter-Adaptive ranked \textbf{\#2} on the public
      RouterArena leaderboard~\citep{lu2025routerarena}, with arena
      score $72.08$ and $75.54\%$ accuracy at \$1.00 per 1K queries.
\end{enumerate}

\section{Methodology}
\label{sec:method}

OrcaRouter treats LLM routing as a \emph{contextual bandit}. At each
round, the router observes a request, selects one model from the
currently active model pool, and receives a scalar reward that
reflects response quality, generation cost, and operational penalties.
The active pool is fixed within a deployment or evaluation interval,
but can be reconfigured between intervals as candidate models are
added, removed, or updated. The policy is initialized offline from
full-information feedback and then deployed either in a frozen
exploitation mode or in a continued-update mode that learns from
partial-information bandit feedback.

\subsection{Problem Formulation}
Let $\mathcal{A}=\{1,\dots,K\}$ denote the model pool of candidates. At
round $t$, the router receives a request $q_t$, encodes it into a context
vector $x_t\in\mathbb{R}^d$, selects an arm $a_t\in\mathcal{A}$, and
observes a scalar reward $r_t\in\mathbb{R}$. The objective is to
maximize the cumulative reward $\sum_{t=1}^{T} r_t$.

\subsection{Context Encoding}
The context vector concatenates two complementary representations.
First, we use a 40-dimensional handcrafted lexical vector
$x^{\text{lex}}_t$ that captures request intent and shape, including
length features, task indicators, formatting cues, and budget-related
signals. Second, each request $q_t$ is
encoded with the frozen \textbf{all-MiniLM-L6-v2}\footnote{\url{https://huggingface.co/sentence-transformers/all-MiniLM-L6-v2}}
model~\citep{reimers2019sentence,wang2020minilm}
into a 384-dimensional L2-normalized embedding $e_t = \phi(q_t)$. The
augmented context is $x^{\text{aug}}_t = [x^{\text{lex}}_t;\, e_t]
\in \mathbb{R}^{424}$.

\subsection{LinUCB Policy}\label{sec:linucb_policy}
We use LinUCB~\citep{li2010contextual,chu2011contextual} to solve the routing task. Each
arm $a$ maintains a regularized Gram matrix
$A_a \in \mathbb{R}^{d \times d}$ and a reward-weighted feature
accumulator $b_a \in \mathbb{R}^d$, initialized as
$A_{a,0} = \lambda I$ and $b_{a,0} = 0$. The score for arm $a$ at
round $t$ is
\begin{equation}
  s_a(x_t) = x_t^\top \hat{\theta}_a
             + \alpha \sqrt{x_t^\top A_a^{-1} x_t},
  \qquad \hat{\theta}_a = A_a^{-1} b_a,
  \label{eq:linucb}
\end{equation}
where $\alpha$ controls exploration. Unless otherwise stated, we use
$\alpha=1.0$ for UCB-based partial-information variants. After
observing reward $r_t$ on the selected arm, the per-arm state updates
as
\begin{equation}
  A_{a_t} \leftarrow A_{a_t} + x_t x_t^\top,
  \qquad
  b_{a_t} \leftarrow b_{a_t} + r_t x_t,
  \label{eq:linucb-update}
\end{equation}
and we maintain $A_a^{-1}$ incrementally via Sherman--Morrison rank-1
updates.

\paragraph{Alternative exploration strategies.}
The default UCB rule is one of several exploration schemes we evaluate
in Section~\ref{sec:exp-online}: (i)~\emph{Linear Thompson Sampling} (LinTS)~\citep{agrawal2013thompson},
sampling $\tilde\theta_a \sim \mathcal{N}(\hat\theta_a,\, v^2 A_a^{-1})$
per step and picking $\arg\max_a \tilde\theta_a^\top x_t$;
(ii)~\emph{$\epsilon$-greedy LinUCB}, picking a uniformly random arm
with probability $\epsilon$ and the top-scoring UCB arm otherwise; and
(iii)~\emph{round-robin warmup + UCB} (RR+UCB), which forces each arm
to be selected at least $n_{\text{RR}}$ times before the UCB head
takes over. We evaluate $n_{\text{RR}}\in\{30,100,300\}$ and the noise
scale $v$ for LinTS in Section~\ref{sec:exp-online}. These variants
mitigate single-arm collapse when bandit feedback is sparse relative
to the feature dimension.

\subsection{Reward Design}
Each served request yields a scalar reward combining four quantities: a quality
score $g_t \in [0,1]$, a normalized per-call cost $\tilde c_t \in
[0,1]$, a normalized latency $\tilde\ell_t \in [0,1]$, and an
operational penalty $p_t$ for discrete failures such as rate-limiting
or malformed output:
\begin{equation}
  r_t = w_q \, g_t
        - w_c \, \tilde c_t
        - w_\ell \, \tilde\ell_t
        - p_t,
  \label{eq:reward}
\end{equation}
with $p_t = w_{\text{rl}}\,\mathbb{1}[\text{rate-limited}] +
w_{\text{fmt}}\,\mathbb{1}[\text{format failed}]$. The cost and latency
terms are clipped against fixed per-call budget scales,
\begin{equation}
  \tilde c_t = \min\!\left(1,\, \frac{c_t}{c_{\max}}\right),
  \qquad
  \tilde\ell_t = \min\!\left(1,\, \frac{\ell_t}{\ell_{\max}}\right),
  \label{eq:costnorm}
\end{equation}
where $c_t$ is per-call USD cost, $\ell_t$ is wall-clock latency in milliseconds, and
$c_{\max}, \ell_{\max}$ are calibration constants set per deployment.
The quality score $g_t$ is supplied by a deployment-specific feedback
channel: a benchmark grader during evaluation, or an operator-defined
validator, model-based judge, or user-feedback signal in production.
Default weights are $w_q{=}1.0$, $w_c{=}0.4$, $w_\ell{=}0.3$,
$w_{\text{rl}}{=}0.5$, and $w_{\text{fmt}}{=}0.3$.

\subsection{Hybrid Offline--Online Learning Protocol}
\label{sec:hybrid}

\paragraph{Offline warmup with full-information feedback.}
We warm up the bandit on a curated set of routing prompts
$\mathcal{D}_{\text{train}}=\{q_1,\dots,q_M\}$ for which every arm's
reward is known. For each query, the LinUCB update
(Eq.~\ref{eq:linucb-update}) is applied to \emph{all} $K$ arms
simultaneously, so each arm's state evolves under full-information
feedback rather than the single-arm signal seen at deployment. Because
every arm observes every query, the sequential updates admit a
closed-form equivalent,
\begin{equation}
  \hat\theta_a
  = (X^\top X + \lambda I)^{-1} X^\top \mathbf{r}_a,
\end{equation}
where $X\in\mathbb{R}^{M\times d}$ stacks the warmup context vectors,
$\mathbf{r}_a\in\mathbb{R}^M$ is the vector of arm-$a$ rewards across
the warmup queries, and $\lambda>0$ is the ridge regularization
strength. The result is a calibrated
initialization for each $\hat\theta_a$ at the start of deployment.

\paragraph{Online deployment.}
At deployment time, the warmed-up router operates in one of two modes.
In the \emph{frozen exploitation} mode, $\hat\theta$ is held fixed and
the exploration bonus is disabled (equivalently $\alpha{=}0$); the
router selects $\hat a_t = \arg\max_a \hat\theta_a^\top x_t$ and fires
a single model call, observing only $r_{\hat a_t, t}$. In the
\emph{continued-update} mode, the router selects arms using the full
LinUCB score (Eq.~\ref{eq:linucb}) and updates the selected arm's
state via Eq.~\ref{eq:linucb-update} after each served request (partial-information feedback). The continued-update mode can adapt to drift, but may introduce
additional exploration variance during deployment.

\section{Experiments}
\label{sec:experiments}

We evaluate OrcaRouter on RouterArena~\citep{lu2025routerarena}, a
public benchmark for comparing LLM routers across accuracy, cost, and
other deployment-relevant metrics. We focus on the Arena score, which
summarizes the accuracy--cost trade-off. RouterArena contains
$8{,}400$ evaluation queries drawn from diverse domains and task
families.

For all experiments, we use a fixed ten-model candidate pool
($K{=}10$): Claude Haiku 4.5, Claude Sonnet 4, Gemini 2.5 Flash,
Gemini 2.5 Flash-Lite, GPT-4o-mini, GPT-5-mini, DeepSeek-chat,
DeepSeek-reasoner, Qwen3 235B Instruct, and Qwen3 30B Instruct.
The pool is fixed throughout evaluation and reflects the model
availability and pricing snapshot used at submission time.

\paragraph{Metric.}
RouterArena reports realized accuracy, average cost per 1K queries,
and a log-cost Arena score. We compute the normalized score as
\begin{equation}
  S_{\mathrm{norm}} =
  \frac{(1+\beta)\,a\,C}{\beta\,a + C},
  \qquad \beta=0.1,
\end{equation}
where $a\in[0,1]$ is accuracy and $C\in[0,1]$ is the normalized
log-cost score. Following the RouterArena leaderboard convention, we
report
\[
  S_{\mathrm{arena}} = 100 S_{\mathrm{norm}},
\]
so an entry with $S_{\mathrm{norm}}=0.7208$ is reported as Arena
score $72.08$. For brevity, we write $S$ for $S_{\mathrm{arena}}$
throughout the rest of the experiments.

\paragraph{Baselines and oracle.}
We compare against the best constant routing policy in our model pool.
In this setting, the strongest single-arm baseline is
\emph{Always-DSC}, which always routes to DeepSeek-chat and obtains
$70.27\%$ accuracy, \$0.10 per 1K queries, and $S{=}70.31$. We also
report a non-deployable oracle upper bound for the fixed model pool:
for each RouterArena query, the oracle selects the best-scoring model
according to the fully observed reward matrix. This oracle obtains $S{=}80.72$, indicating that OrcaRouter-Adaptive
remains below the best achievable score within the fixed model pool.

\paragraph{Training matrices.}
Except for the explicitly marked full-information diagnostic fits,
RouterArena prompts are used only for evaluation. Training uses a curated warmup set together with prompts
drawn from RouterBench~\citep{hu2024routerbench}. The curated warmup
set is an internal prompt set used only for
initialization. After filtering out $1{,}135$ near-duplicates whose
MiniLM cosine similarity to any RouterArena prompt exceeds $0.85$, we retain $5{,}000$ RouterBench prompts and evaluate all ten candidate
models on them, yielding a $5{,}000 \times 10$ reward matrix. We
initialize the per-arm ridge states using the closed-form ridge
solution described in Section~\ref{sec:hybrid}. In RouterArena experiments, we compute rewards from the benchmark
quality score and provider pricing. We treat latency and operational
penalties as zero unless we explicitly measure them. We separately report RouterArena
in-distribution full-information results as diagnostic upper bounds.

\subsection{RouterArena Results}
\label{sec:exp-online}

We evaluate the proposed routing strategies on RouterArena using the
same fixed ten-model pool. Table~\ref{tab:online} reports the main
results, including the best constant baseline, partial-information
bandit variants, full-information diagnostic fits, the deployed
submission, and the oracle. All scores in the table are evaluated on
the full RouterArena evaluation set of $8{,}400$ prompts.

\begin{table}[h]
\centering
\small
\caption{Routing strategies on our fixed model pool. Arena score $S$ is
reported on the leaderboard 0--100 scale. Partial-information bandit
rows report mean $\pm$ standard deviation over three seeds.}
\label{tab:online}
\begin{tabular}{@{}lccc@{}}
\toprule
Strategy & Arena $S$ & Top arm \% & \# arms used \\
\midrule
Always-DSC & 70.31 & 100\% & 1 \\
\midrule
Vanilla LinUCB ($\alpha{=}1$) & $69.81 \pm 0.40$ & 15.5\% & 10 \\
LinTS ($v{=}0.3$) & $70.00 \pm 0.42$ & 72.8\% & ${\approx}2$ \\
$\epsilon$-greedy LinUCB & $70.71 \pm 0.25$ & 20.0\% & 10 \\
RR+UCB & $70.74 \pm 0.08$ & 18.2\% & 10 \\
\midrule
Full-info batch fit (default $\lambda$) & 73.54 & 26.0\% & 10 \\
Full-info batch fit (tuned) & 74.05 & 31.8\% & 10 \\
\midrule
\textbf{Submission} & \textbf{72.08} & \textbf{19.7\%} & \textbf{10} \\
Oracle & 80.72 & 39.4\% & 10 \\
\bottomrule
\end{tabular}
\end{table}

\paragraph{Partial-information learning.}
The four partial-information variants start without offline warmup
($A_{a,0}{=}\lambda I$, $b_{a,0}{=}0$). They are trained on
RouterBench prompts for which we collected rewards over the ten-model
pool. During training, prompts are replayed under a bandit-feedback
protocol: at each step, the router selects one model, observes only
that model's reward, and updates only the selected arm. The learned
policy is then evaluated on the RouterArena evaluation set. Vanilla
LinUCB uses only the standard UCB exploration bonus, while LinTS,
$\epsilon$-greedy LinUCB, and RR+UCB add explicit exploration
mechanisms to mitigate under-exploration in the cold-start
partial-information regime.

\paragraph{Full-information diagnostic fits.}
The \emph{Full-info batch fit} rows replace bandit feedback with a
closed-form ridge fit using RouterArena reward matrices, where every
arm's reward is observed for every prompt during fitting. These rows
are diagnostic rather than deployable: they estimate what this model
class can achieve when in-distribution full-information feedback is
available. The tuned row uses $\lambda{=}0.1$ and $w_c{=}0.35$, selected in a
separate hyperparameter tuning step on a held-out RouterArena fold and
then reported as a full-set diagnostic fit. They should be
read as diagnostic reference points, not as leaderboard submissions.

\paragraph{Deployed submission.}
OrcaRouter-Adaptive is initialized with full-information warmup on a
separate curated routing-prompt set, then updated using the collected
RouterBench reward matrix under partial-information bandit feedback.
The deployed submission is different from the diagnostic fits. It is
not fitted on RouterArena and is evaluated frozen, with no policy
updates from RouterArena prompts. At the time of our RouterArena submission, this configuration ranked
\textbf{\#2} with accuracy $75.54\%$ at \$1.00 per 1K queries and
arena score $\mathbf{S{=}72.08}$.

\subsection{Analysis}
\label{sec:exp-analysis}

\paragraph{Pick diversity.}
A common concern in cost-aware routing is collapse to a single cheap
model. Under the best diagnostic full-information configuration
($S{=}74.05$), the most frequently selected arm, DeepSeek-chat, handles
approximately $32\%$ of queries while all ten arms remain active.
Replacing the embedding-augmented scorer with a bias-only
non-contextual ridge model, which has no request-level features and
learns only global arm preferences, collapses the router to always
selecting DeepSeek-chat. This is mathematically equivalent to a
hard-coded single-model policy. The contrast shows that the contextual
embedding carries genuine per-query routing signal.

\paragraph{Robustness.}
RouterArena also evaluates routers on paraphrased prompts to measure
pick stability. The submitted system obtains a RouterArena robustness
score of $22.62$, indicating that robustness is the main area for improvement in the
current system. We attribute this behavior largely to narrow score
margins between the top-ranked arms. Small embedding shifts caused by
paraphrasing can change the arm with the highest score, even when the
underlying request intent is similar. We define pick-flip rate as the
fraction of paraphrased prompt pairs for which the selected arm differs
from the original prompt.

A simple margin-based tie-breaker reduces this instability. When the top two arms score within $\epsilon{=}0.02$, the router selects the arm with the higher mean reward on the training data. This
rule reduces the pick-flip rate from $16.7\%$ to $2.4\%$, while
reducing $S$ by roughly $0.86$ points. We treat this rule as a
deployment switch for workloads that prioritize routing stability over
the highest possible Arena score.

\section{Conclusion}
\label{sec:conclusion}

We presented \textbf{OrcaRouter}, a production-oriented LLM router
that formulates query-level LLM routing as a contextual bandit with a
configurable multi-model action space. OrcaRouter combines
full-information offline initialization with optional
partial-information online updates, allowing the routing policy to
adapt from feedback observed for the selected model.

At the time of our RouterArena submission (May 20, 2026),
OrcaRouter-Adaptive ranked \textbf{\#2} on the public RouterArena
leaderboard with arena score $S{=}72.08$, achieving $75.54\%$ accuracy
at \$1.00 per 1K queries. Our experiments
support a hybrid view of production-oriented LLM routing.
Full-information offline warmup provides a strong initialization, while
bandit feedback enables continual adaptation under realistic serving
constraints. Future work should focus on improving robustness under
paraphrase-induced embedding shifts, reducing the sample complexity of
online adaptation, and expanding the training prompt mixture to better
cover task families where the model pool offers meaningful
quality--cost trade-offs.

\bibliographystyle{unsrtnat}
\bibliography{references}

\end{document}